# Data Augmentation for Cognitive Behavioral Therapy: Leveraging ERNIE Language Models using Artificial Intelligence

**Bosubabu Sambana**
*Assistant Professor,
Department of Computer Science and Engineering ( Data Science )
School of Computing,
Mohan Babu University, Tirupati, India.*
bosubabu.s@mbu.asia

**Kondreddygari Archana**
*UG Scholar,
Department of Information Technology
Sree Vidyanikethan Engineering
College Tirupati,India.*
kondreddygariarchana@gmail.com

**Suram Indhra Sena Reddy**
*UG Scholar,
Department of Information Technology
Sree Vidyanikethan Engineering
College Tirupati,India.*
suramindhrasenareddy@gmail.com

**Shaik Meethaigar Jameer Basha**
*UG Scholar,
Department of Information Technology
Sree Vidyanikethan Engineering
College Tirupati,India.*
shaikjameer1161@gmail.com

**Shaik Karishma**
*UG Scholar,
Department of Information Technology
Sree Vidyanikethan Engineering
College Tirupati,India.*
karishmashaik56@gmail.com

*Abstract*—**Cognitive Behavioral Therapy (CBT) is a proven approach for addressing the irrational thought patterns associated with mental health disorders, but its effectiveness relies on accurately identifying cognitive pathways to provide targeted treatment. In today's digital age, individuals often express negative emotions on social media, where they may reveal cognitive distortions, and in severe cases, exhibit suicidal tendencies. However, there is a significant gap in methodologies designed to analyze these cognitive pathways, which could be critical for psychotherapists aiming to deliver timely and effective interventions in online environments. Cognitive Behavioral Therapy (CBT) framework leveraging acceptance, commitment and data augmentation to categorize and address both textual and visual content as positive or negative. Specifically, the system employs BERT, RoBERTa for Sentiment Analysis and T5, PEGASUS for Text Summarization, mT5 for Text Translation in Multiple Languages focusing on detecting negative emotions and cognitive distortions within social media data. While existing models are primarily designed to identify negative thoughts, the proposed system goes beyond this by predicting additional negative side effects and other potential mental health disorders likes Phobias, Eating Disorders. This enhancement allows for a more comprehensive understanding and intervention strategy, offering psychotherapists a powerful tool for early detection and treatment of various psychological issues.**

## I. INTRODUCTION

The brain is an intricate organ that controls thinking, feelings, and actions. In the framework of Cognitive Behavioral Therapy (CBT), brain activity highlights the connection between mental processes, such as interpretation and judgment, and managing emotions. CBT aims to identify and modify unhelpful thought patterns to reshape neural connections and encourage healthier habits. Data augmentation is a pivotal strategy in enhancing the performance and robustness of machine learning systems, especially in domains like cognitive behavioral therapy (CBT), where high-quality labeled data is often scarce. In the context of Acceptance and Commitment Therapy (ACT), data augmentation can significantly improve the identification, analysis, and classification of cognitive pathways by expanding and diversifying datasets. This helps in training more effective models for understanding and responding to psychological patterns. Incorporating advanced natural language processing (NLP) models like BERT and T5 further refines these processes, enabling precise interventions. This Bidirectional Encoder Representations for Transformers used for Sentiment analysis in Classifying the text input by the User. Text summarization uses PEGASUS for Simplification of the data in User content for easy understanding.

Data augmentation in Cognitive Behavioral Therapy (CBT) focuses on expanding and diversifying datasets to capture nuanced therapeutic contexts, such as thoughts, emotions, and behaviors. Using advanced models like RoBERTa and PEGASUS, data augmentation rapidly generates varied and contextually relevant text, significantly enhancing the ability to recognize cognitive patterns, classify psychological states, and simulate personalized therapy interventions. Depression is a form critical situation that users go threw in daily life because of situations at work.

Individuals who have experienced abuse, significant losses, or other adverse events are more prone to developing depression. Challenges in school or work environments can also contribute to its onset. Social media platforms are frequently used by young people to express suicidal thoughts and emotions. Reliability can be achieved by conducting user studies, and employing cross validation techniques to avoid Overfitting. The recognition process can be done by using sequence-to-sequence models for intent recognition and cognitive pathway extraction. However, there is limited understanding of how social media can be leveraged for suicide prevention. This study aimed to systematically review existing evidence on the current use of social media as a tool for preventing suicide.

*Core Aspects of Data Augmentation:*
- *Text Transformations:* Restating sentences using different words or sentence structures while maintaining the original meaning.
- *Synonym Replacement:* Substituting certain words with their synonyms to generate alternative expressions. It do not change the meaning but alternative word is used.
- *Back-Translation:* Translating a sentence to another language and back to its original language to create variations while preserving meaning.
- *Character-level Noise:* Introducing typos or spelling errors to simulate real-world text.
- *Word Dropout:* Randomly removing words to simulate incomplete sentences.
- *Shuffling Words:* Rearranging words in a sentence while retaining grammatical correctness.
- *Synthetic Data Generation:* Using models like GPT, T5, or BERT to create new text data relevant to the domain.

## II. LITERATURE SURVEY

The World Health Organization emphasizes that depression arises from a complex interplay of social, psychological, and biological factors, including adverse life events such as abuse and significant losses. Psychological interventions like Cognitive Behavioral Therapy (CBT) have been validated as effective treatments, addressing the negative thought patterns contributing to depression's persistence [1].Huang et al. conducted a large-scale study in China, revealing a high prevalence of mental disorders, including depression. This emphasizes the urgent need for scalable, evidence-based interventions like CBT to address mental health challenges at the population level [2].

P.Cuijpers et al. conducted a meta-analysis comparing CBT to other treatments such as pharmacotherapy and alternative psychotherapies. The study, which encompassed 409 trials and over 52,000 patients, affirmed CBT's superior efficacy and its foundational role in treating depression [3][4].Ellis and Dryden presented Rational Emotive Behavior Therapy (REBT), which laid the groundwork for CBT by focusing on identifying and altering irrational beliefs. This seminal work continues to guide CBT's theoretical and practical applications [5].

William et al. explored the potential of leveraging BERT-based extractive summarization techniques for detecting depression on social media, demonstrating significant advancements in identifying mental health conditions through natural language processing (NLP). Their study highlights the role of AI in analyzing user-generated content to provide scalable and efficient tools for mental health detection, aligning with modern therapeutic approaches such as CBT [6].

Andrews et al. examined the challenges in categorizing anxiety and depressive disorders, proposing refined classification methods to address overlaps and diagnostic ambiguities. Their work underscores the importance of precise diagnostic frameworks in enhancing therapeutic outcomes, offering a foundation for integrating technologies like AI and CBT to better tailor interventions for individual patient needs. [7].

Rani et al. highlighted the development of a mental health chatbot leveraging NLP and AI to deliver Cognitive Behavioral Therapy (CBT) and facilitate remote health monitoring. Their work represents a significant step toward personalized and accessible mental health care, showcasing how AI-driven solutions can complement traditional therapeutic methods by offering real-time support and improving patient engagement. [8].

Aragon et al. developed Disorder BERT, a domain-adaptive model for detecting mental health disorders in social media posts. Similarly, Zhai et al. (2024) introduced Chinese Mental BERT, tailored for analyzing mental health in Chinese-language texts. Both models demonstrate the power of specialized AI in addressing cultural and linguistic nuances in mental health care[9].He et al. discussed the potential for psychological generalist AI, capable of addressing diverse mental health scenarios. Qi et al. demonstrated the use of supervised learning and LLMs for identifying cognitive distortions and suicidal risks [10][11].

Sun et al. introduced ERNIE 3.0, a large-scale knowledge-enhanced pre training model. Its integration into CBT can enhance the identification of complex cognitive pathways and patterns, paving the way for more effective interventions [12] Meghrajani et al. conducted an in-depth analysis of mental health challenges in India, emphasizing the historical and contemporary significance of mental asylums in addressing these issues. Their findings provide critical insights into systemic barriers and potential frameworks for improving mental health care accessibility, laying a foundation for integrating modern approaches like CBT and AI-driven tools. [13].

Parviainen and Rantala examined the rise of chatbot-based consultations in health care during the 2020s, offering an ethical perspective on this trend. Their work underscores the potential of automated systems to enhance mental health service delivery, while also addressing concerns related to patient trust and data security, making these systems a complement to traditional therapies such as CBT. [14].

## III. EXISTING SYSTEM

Existing System represents a system framework designed for understanding and analyzing user emotions to assist therapists in providing effective and timely therapy. The process begins with considering the user's input thoughts using Hierarchical Text Classification, which categorizes the text into different predefined categories, providing a structured analysis.

Text Summarization process considers the weighted content, extracting the most occurred keywords for further suggestions. Finally, Text Conversion ensures the data is transformed into a user understandable format preparing it for deeper sentimental and emotional analysis (Fig .1).

The ABCD Model at the heart of this system involve the core aspects of Rational-Emotive Behavior Therapy to identify the Activating Event, Belief, Consequence, and Disputation. To achieve this, Convolutional Neural Networks are employed for encoding the hierarchical and spatial relationships within the text data, while Recurrent Neural Networks, particularly Long Short-Term Memory networks, are utilized for decoding sequential patterns, such as emotional transitions and cognitive pathways.

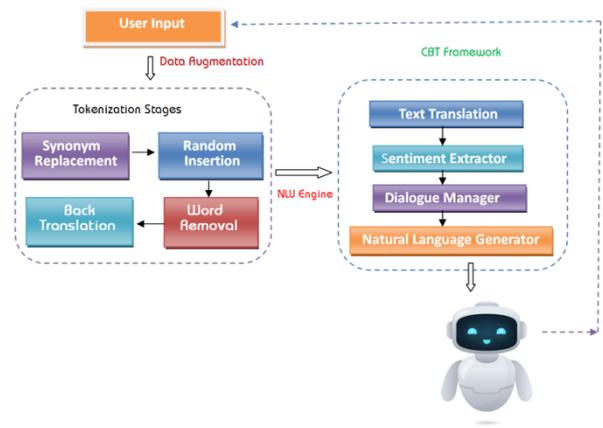

Fig 2. Proposed Architecture Working Flow

User Input: Represents text provided by the user, which could include their emotions, concerns, or phobias specific issues regarding daily life activities. This input forms the foundation for analysis and processing.

*Data Augmentation:* Augments the input to make the model robust and adaptable by generating different variations of the actual text. Ensures better training for the system to handle different phrasing and contexts. It involve tokenization techniques like synonym replacement, Random insertion, Back Translation, Word removal. It increases corpus extracting quicker based on user requirement.

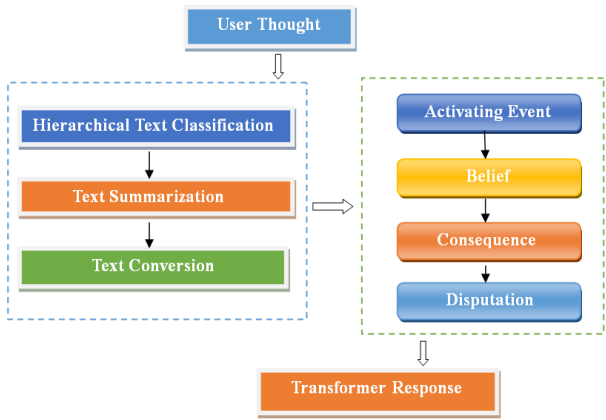

Fig 1. Internal Operational Working Mechanism

Further, a GPT-3.0-based transformer model integrates these processes to analyze the user's mental state. This language model leverages its deep learning capabilities, to classify the user's mood and emotional patterns. By detecting subtle linguistic and contextual cues, GPT-3.0 acts as a bridge between raw input data and actionable insights for therapists. Its output includes a synthesized analysis of the user's thoughts, highlighting specific cognitive distortions or negative thought patterns [15].

## IV. RESEACH WORK

The Research Work is an AI-based Posibot designed to provide Cognitive Behavioral Therapy support. The system improves on previous models by integrating data augmentation, natural language understanding (NLU), and specific modules for sentiment analysis, dialog management, and therapeutic suggestions (Fig 2.). It also addresses phobias by offering actionable suggestions and exercises to promote healthy living.

*NLU Engine:* Processes the augmented data to extract intent, emotions, and entities from the user's input that help user to know they are going through, Acts as a link between the input text and the CBT framework.

*CBT Framework:*

- *Text Translation:* Converts user input into a structured format for easier processing. Supports multilingual interactions.

- *Sentiment Extractor:* Analyzes the user's emotions to determine their state of mind and identify negative thought patterns.

- *Dialog Manager:* Facilitates meaningful and context-aware conversations with the user. It ensures continuity and provides relevant responses.

- *Natural Language Generator (NLG)***:** Generates empathetic and actionable responses, making the AI posibot more reliable and effective to provide suggestions to user.

Provides tailored suggestions for improving mental health, like guided exercises, providing encouraging suggestions and mechanisms. If the input indicates a phobia, the Posibot suggests step-by-step exposure therapy exercises or relaxation techniques to help users overcome their fears and lead healthier lives.

### A. ALGORITHM

The Augmented Sentiment-Aware Response Generation algorithm processes user Data to create personalized, sentimental responses. It starts by generating augmented variations of the input text using methods like synonym replacement, back-translation, and paraphrasing. For early detection process a subtle negative emotions indicator refers to technique used to detect and measure low intensity or nuanced negative emotions in CBT. These various terms are then translated using the mT5 model to enhance multi-linguistic diversity. Sentiment analysis is done using pre-trained models like BERT or RoBERTa to detect the emotional tone of the text. The text is summarized using models like T5 or Pegasus to grab key information. Finally, the sentiment and summary are combined to provide a response tailored to the user's emotional mode.

Step-1: Let input text T, INPUT=T that is entered by the User.

$$Input = T_{User}$$

Step-2: Input text Augmentation, Let input text T generate augmented versions by applying augmentation techniques.

$$T_{augmented} = \{T_{augmented1}, T_{augmented2}, \ldots, T_{augmentedN}\}$$

$$T_{augmented} = \{T_{synonym}, T_{back\_translate}, T_{paraphrase}\}$$

Step-3: Apply mT5 for translating the augmented text into another language and back to the original language.

$$T_{augmented\_translated} = mT5(T_{augmented})$$

Step-4: For each translated augmented text, apply sentiment classification using a pre-trained BERT or RoBERTa model.

$$BERT(T) = p(x_t \mid x_1, \ldots x_{t-1}, x_{t+1}, \ldots, t_n)$$

$$\hat{Y}_{sentiment} = Softmax(W \cdot E_{augmented\_translated} + b)$$

Step-5: Apply text summarization to the translated text using T5 or Pegasus.

$$T_{summary} = T5(T_{augmented\_translated})$$

Step-6: Based on the sentiment prediction, generate a response.

$$Response = f(\hat{Y}_{sentiment}t, T_{summary})$$

### B. Implementation

The Suicide Watch Dataset, Sentiment Analysis for Mental Health Dataset, and Student Depression Dataset provide essential requirements for designing a comprehensive mental health Posibot. These data contribute to different facets of understanding and categorizing mental health, enabling the creation of an supportive, AI-driven system that classify the different needs of users experiencing mental health issues. These data provide a strong foundation for the designing of AI-powered CBT systems that can result tailored, effective support for users dealing with mental health issues. The combination of variety datasets ensures that the CBT model is exposed to a broad range of user experiences and emotional expressions, which increases its accuracy and performance to classify and respond to different mental health conditions.

*Suicide and Depression Detection Dataset:* The Suicide Watch Dataset is a crucial asset for mental health research, particularly in analyzing suicidal thoughts expressed online. This dataset includes a vast collection of posts discussing suicide, making it highly beneficial for exploring mental health patterns and associated risk factors. It comprises text entries sourced from Kaggle, sorted into categories (e.g., suicidal or non-suicidal) to facilitate tasks such as sentiment analysis, text categorization, and natural language processing (NLP).The dataset contains 5 attributes with 348,000 data entries, covering key elements like the content of posts, timestamps, authors, subreddits, and their respective labels for classification. The dependent variable, termed "class," has 2 distinct categories representing suicidal and non-suicidal posts, enabling tasks involving supervised learning. The Suicide Watch Dataset plays a pivotal role in enhancing AI-powered approaches for mental health support, serving as a cornerstone for systems designed to identify and address suicide risks in a timely manner [15].

*Sentiment Analysis Dataset:* The Sentiment Analysis for Mental Health Dataset is a vital resource for examining mental health sentiments and their related patterns. It emphasizes analyzing the emotional tone of text statements associated with mental health, making it highly applicable for activities such as sentiment evaluation, categorization, and natural language processing (NLP).This dataset contains 53,074 records with 3 main columns. A unique identifier allocated to each entry. Text content reflecting mental health-related expressions. The target variable that classifies mental health status, including Normal, Depression, and Other. This dataset is crucial for developing AI-driven models that aim to assess and forecast mental health conditions based on text data. By facilitating sophisticated machine learning and NLP applications, it supports the creation of advanced tools for mental health assessment, intervention, and treatment. This would help in classifying type of sentiment that user going through in daily life activities.

*Student Depression Dataset:* The Student Depression Dataset is a comprehensive resource designed to explore depression levels among students, capturing various demographic and psychological factors. This dataset provides valuable insights for mental health research, particularly in educational settings. By leveraging its diverse features like age, gender, academic year, CGPA, marital status, financial situation, and depression levels, it provides a multidimensional view of factors influencing student mental health. This allows the creation of machine learning models that classify user input into predefined categories, such as "low," "moderate," or "severe" depression levels. The dataset comprises approximately 1,015 responses. Each column reflects diverse categorical or numerical data points tailored to understanding mental health in students. Aggregated column descriptions can provide insights into factors correlating with student depression. A breakdown of descriptive responses shows the dataset contains approximately 5,000+wordsacross categorical labels and descriptions. This dataset offers a solid foundation for analyzing the impact of demographic and psychological factors on mental health, aiding in creating interventions and support systems for students.

## V. RESULT ANALYSIS

The Results and analysis focuses on how the model process various emotional states, their coordination with age and gender, and how these various insights can improve therapeutic interventions. The results highlight the need for a scalable, data-driven CBT model that can accommodate the emotional and demographic diversity observed in the dataset. By recognizing the sentimental variances based on age and gender, the model can be optimized to offer more personalized, effective outcomes. Additionally, the analysis of sentence length diversity ensures that the Bot can handle a broad distribution of user inputs, making it a more robust tool for real-time therapy.

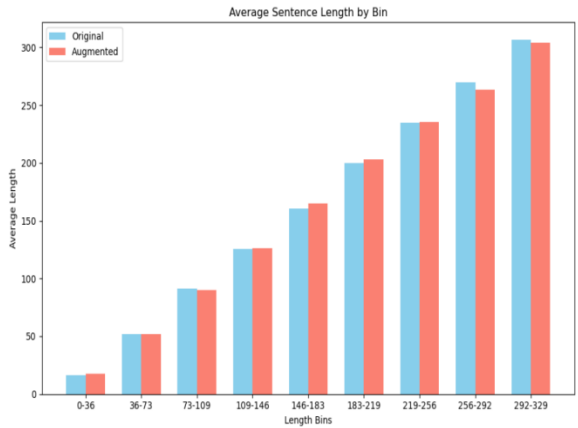

Fig 3. Comparative Analysis of Sentence Length Distribution

The x-axis of the graph represents length bins, which group sentences based on their length ranges (e.g., 0–36, 36–73, and so on). These bins provide a clear segmentation of input data, allowing for the analysis of how sentence lengths vary before and after augmentation (Fig.3). It is generated by collecting original and augmented sentences then performed a count operation on both datasets, and then calculating the average sentence length separately for both the original and Augmented datasets using pandas. For instance, the lower bins such as 0–36 represent very short sentences, while higher bins such as 109–146 and above capture longer and more detailed sentences.

The y-axis indicates the average sentence length within each bin, measuring the count of sentences that fall into these ranges. This visualization highlights the changes brought about by augmentation and ensures a more balanced dataset for training. This helps in generating helpful Solutions by the Model. By employing data augmentation, the dataset achieves greater diversity in sentence lengths, enabling the chatbot to better understand, classify, and respond to a wide variety of user inputs

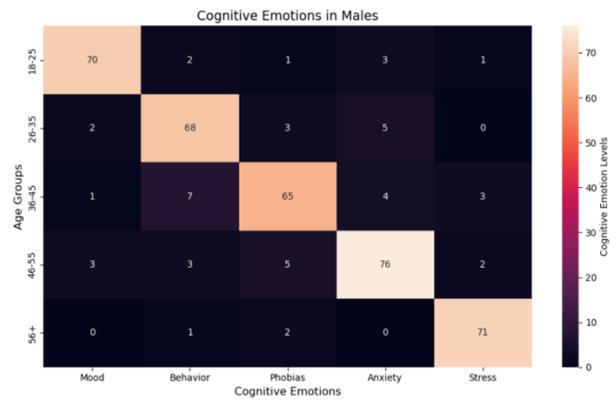

Fig 4. Distribution of Cognitive Emotions Across Age Groups in Males

The map presented above visualizes the distribution of cognitive emotion levels among males across different age groups. The x-axis represents various cognitive emotions, including Mood, Behavior, Phobias, Anxiety, and Stress, which are key focus areas in Cognitive Behavioral Therapy (CBT) (Fig. 4.). The y-axis categorizes males into different age groups: 18–25, 26–35, 36–45, 46–55, and 56+. This segmentation allows for a detailed analysis of how cognitive emotions vary by age, helping to identify patterns specific to certain age ranges.

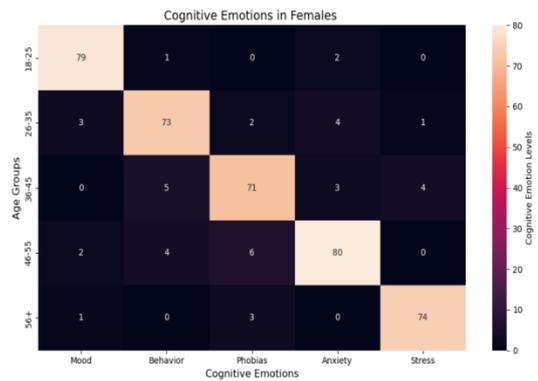

Fig.5 Distribution of Cognitive Emotions Across Age Groups in Females

The color intensity in the map represents the cognitive emotion levels, as indicated by the color bar on the right. Lighter shades (closer to white) indicate higher levels of a particular cognitive emotion in that age group, while darker shades (closer to black) signify lower levels. The numerical values within each cell provide exact measurements for better clarity [15].

This graph is crucial for CBT applications as it sheds light on how emotional patterns differ among females of various age groups. Compared to the male map, significant differences can be observed. Females exhibit a higher prevalence of mood-related issues in the 18–25 age group, while anxiety levels peak sharply in the 46–55 group, much higher than in males. Stress levels in females are also noticeably higher in the 56+ group, whereas males show a more evenly distributed stress level across age groups (Fig.5). Additionally, behavioral issues appear to be more prominent in females aged 26–35, contrasting with their lower presence in males. These differences highlight the need for gender-specific considerations in CBT-based therapeutic designs.

## VI. CONCLUSION

Cognitive Behavioral Therapy (CBT) has demonstrated remarkable potential in addressing mental health challenges by identifying and restructuring negative thought patterns to foster healthier behaviors and emotional well-being. The development of a Posibot powered by AI and NLP technologies can significantly enhance the delivery of CBT by classifying, categorizing, and analyzing emotions, behaviors, and moods of individuals in real-time. Ensuring the security of patient data is a challenge, domain specific datasets limited in CBT. A Posibot designed with AI and NLP technologies can revolutionize CBT delivery by analyzing user input to classify emotions, behaviors, and moods. The Posibot utilizes advanced algorithms to categorize user inputs into predefined categories, such as sadness, anxiety, anger, or positive emotions, and generates contextually appropriate therapeutic responses. Such systems enable users to overcome negative thoughts, foster resilience, and lead positive lives.

Additionally, these tools can account for gender differences in emotional processing, recognizing that men and women often vary in their emotional responses and thought patterns, allowing for a more personalized therapeutic experience. The F1-score is a harmonic mean essential in CBT models to balance false positive and false negatives. The negative sentences count and positive sentences count is used in F1-score to enhance precision. By integrating advanced algorithms, such Posibots not only make mental health care more accessible but also offer scalable and efficient support tailored to individual needs.

## VII. FUTURE SCOPE

The future of Cognitive Behavioral Therapy (CBT) Posiots holds immense potential in transforming mental health care into a more accessible, personalized, and dynamic domain. Advanced AI technologies, such as natural language understanding (NLU) and sentiment analysis, can enable Posibots to detect subtle emotional cues, such as tone and word choice, for more empathetic and precise interventions. These Posibots can be developed to dynamically adapt their therapeutic approaches based on user needs, providing a blend of techniques like cognitive reframing, mindfulness exercises, or relaxation strategies to effectively address negative thought patterns. By understanding user emotions in real time, these Posibots can offer meaningful support for overcoming daily challenges.

Future developments could also integrate CBT Posibots with wearable devices to provide a holistic approach to mental health care. Wearables that monitor heart rate, sleep patterns, and physical activity could supply real-time data, helping the Posibot better understand the user's overall well-being. This integration could enable timely suggestions to mitigate stress or anxiety, while personalized treatment plans, developed using machine learning, could refine therapeutic approaches over time based on user interactions. Additionally, multi-lingual and cross-cultural capabilities could make these Posibots accessible to diverse populations, ensuring that users receive relevant and culturally sensitive mental health support

## References

[1] World Health Organization, "Depressive disorder (depression)," World Health Organization, 2023.

[2] Y. Huang, Y. Wang, H. Wang, Z. Liu, X. Yu, J. Yan, Y. Yu, C. Kou, X. Xu, J. Lu et al., "Prevalence of mental disorders in china: a cross sectional epidemiological study," The Lancet Psychiatry, vol. 6, no. 3, pp. 211–224, 2019.

[3] J. Robinson, G. Cox, E. Bailey, S. Hetrick, M. Rodrigues, S. Fisher, and H. Herrman, "Social media and suicide prevention: a systematic review," Early intervention in psychiatry, vol. 10, no. 2, pp. 103–121, 2016.

[4] P. Cuijpers, C. Miguel, M. Harrer, C. Y. Plessen, M. Ciharova, D. Ebert, and E. Karyotaki, "Cognitive behavior therapy vs. control conditions, other psychotherapies, pharmacotherapies and combined treatment for depression: a comprehensive meta-analysis including 409 trials with 52,702 patients," World Psychiatry, vol. 22, no. 1, pp. 105–115, 2023.

[5] Ellis and W. Dryden, The practice of rational emotive behavior therapy. Springer publishing company, 2007.

[6] D. William, S. Achmad, D. Suhartono and A. P. Gema, "Leveraging BERT with Extractive Summarization for Depression Detection on Social Media," 2022 International Seminar on Intelligent Technology Authorized licensed use limited to: Indian Institute of Technology.

[7] Andrews, G., Anderson, T. M., Slade, T., & Sunderland, M. (2008). Classification of anxiety and depressive disorders: problems and solutions. Depression and anxiety, 25(4), 274-281.

[8] K. Rani, H. Vishnoi and M. Mishra, "A Mental Health Chatbot Delivering Cognitive Behavior Therapy and Remote Health Monitoring Using NLP And AI," 2023 International Conference on Disruptive Technologies (ICDT), Greater Noida, India, 2023, pp. 313- 317, doi: 10.1109/ICDT57929.2023.10150665.

[9] M. Aragon, A. P. L. Monroy, L. Gonzalez, D. E. Losada, and M. Montes,"DisorBERT: A Double Domain Adaptation Model for Detecting Signsof Mental Disorders in Social Media," in Proceedings of the 61st AnnualMeeting of the Association for Computational Linguistics (Volume 1:Long Papers), 2023, pp. 15 305–15 318.

[10] T. He, G. Fu, Y. Yu, F. Wang, J. Li, Q. Zhao, C. Song, H. Qi, D. Luo,H. Zou et al., "Towards a psychological generalist ai: A survey ofcurrent applications of large language models and future prospects,"arXiv preprint arXiv:2312.04578, 2023.

[11] H. Qi, Q. Zhao, C. Song, W. Zhai, D. Luo, S. Liu, Y. J. Yu, F. Wang,H. Zou, B. X. Yang et al., "Evaluating the efficacy of supervised learningvs large language models for identifying cognitive distortions and suicidal risks in Chinese social media," arXiv preprint arXiv:2309.03564,2023.

[12] Y. Sun, S. Wang, S. Feng, S. Ding, C. Pang, J. Shang, J. Liu, X. Chen,Y. Zhao, Y. Lu et al., "ERNIE 3.0: Large-scale knowledge enhancedpre-training for language understanding and generation," arXiv preprintarXiv:2107.02137, 2021.

[13] Meghrajani, V.R., Marathe, M., Sharma, R., Potdukhe, A., Wanjari, M.B., Taksande, A.B., Meghrajani Jr, V.R. and Wanjari, M., " A Comprehensive Analysis of Mental Health Problems in India and the Role of Mental Asylums", Cureus, vol. 15, no. 7, 2023.

[14] Parviainen, J. Rantala, J., " Chatbot breakthrough in the 2020s? An ethical reflection on the trend of automated consultations in health care", Medicine, Health Care and Philosophy, vol.25, no.1, pp.61-71, 2022.of Applied Studies, August, 2013, DOI:10.3886/ICPSR30122.v2